\documentclass[10pt,twocolumn]{article}

\usepackage[T1]{fontenc}
\usepackage[utf8]{inputenc}
\usepackage{microtype}

\usepackage[round]{natbib}
\usepackage{mathtools}
\usepackage{tikz}
\usepackage{multirow}
\usepackage{colortbl}
\usepackage{makecell}
\usepackage{amsfonts}
\usepackage{bbm}
\usepackage{bm}
\usepackage{amsmath}
\usepackage{amsthm}
\usepackage{hyperref}
\usepackage{authblk}

\usepackage[margin=0.8in,columnsep=0.3in]{geometry}
\usepackage[indent=0pt]{parskip}
\usepackage{enumitem}

\newtheoremstyle{myplain}%
  {0.8\baselineskip}% Space above
  {0.8\baselineskip}% Space below
  {\itshape}% Body font
  {}% Indent amount
  {\bfseries}% Theorem head font
  {.}% Punctuation after theorem head
  {0.5em}% Space after theorem head
  {}% Theorem head spec

\theoremstyle{myplain}
\newtheorem{theorem}{Theorem}
\newtheorem{proposition}{Proposition}
\newtheorem{definition}{Definition}

\newcommand{\argmin}{\arg\!\min}

\makeatletter
\renewenvironment{abstract}
  {\centerline{\large\bfseries Abstract}
   \vspace{0.5ex}
   \begin{quote}}
  {\par\end{quote}\vspace{1ex}}
\makeatother

\makeatletter
\renewcommand\paragraph{\@startsection{paragraph}{4}{\z@}
  {1.5ex \@plus 0.5ex \@minus .2ex}
  {-1em}
  {\normalfont\normalsize\bfseries}}
\makeatother

\title{\vspace{-0.5em} Calibrating Conditional Risk}

\author[1]{Andrey Vasilyev}
\author[2]{Yikai Wang}
\author[1]{Xiaocheng Li}
\author[2,*]{Guanting Chen}

\affil[1]{Imperial College Business School, Imperial College London}
\affil[2]{Department of Statistics and Operations Research, UNC-Chapel Hill}

\date{\vspace{0.45em}}

\begin{document}

\maketitle

\def\thefootnote{*}\relax\footnotetext{Corresponding to Guanting Chen (guanting@unc.edu).}

\begin{abstract}
We introduce and study the problem of calibrating conditional risk, which involves estimating the expected loss of a prediction model conditional on input features. We analyze this problem in both classification and regression settings and show that it is fundamentally equivalent to a standard regression task. For classification settings, we further establish a connection between conditional risk calibration and individual/conditional probability calibration, and develop theoretical insights for the performance metric. This reveals that while conditional risk calibration is related to existing uncertainty quantification problems, it remains a distinct and standalone machine learning problem. Empirically, we validate our theoretical findings and demonstrate the practical implications of conditional risk calibration in the learning to defer (L2D) framework. Our systematic experiments provide both qualitative and quantitative assessments, offering guidance for future research in uncertainty-aware decision-making.
\end{abstract}

\section{Introduction}
Understanding and quantifying the uncertainty of machine learning models is essential in real-world decision-making systems, especially in applications involving risk-sensitive domains such as autonomous systems and human-AI interaction \citep{abdar2021review,smith2024uncertainty}. Decision-making systems must not only provide accurate predictions but also assess the reliability of those predictions to inform downstream tasks. Traditional uncertainty quantification methods focus on global or marginal uncertainty estimates. However, many practical applications require a more refined characterization of uncertainty that is specific to individual inputs \citep{yang2015multi, cortes2016learning, madras2018predict}. This motivates the study of conditional risk calibration, where the goal is to estimate the expected loss of a given predictor conditioned on input features. A well-calibrated conditional risk estimator provides an interpretable measure of predictive quality, enabling more robust decision-making in safety-critical and resource-sensitive applications.

Despite its importance, the problem of conditional risk calibration/estimation has received limited formal treatment in the literature. Existing methods that implicitly rely on conditional risk estimation often approach it heuristically \citep{shrivastava2016training,yoo2019learning}, without a rigorous formulation or understanding of its theoretical properties. 

This work provides a comprehensive theoretical and empirical investigation of conditional risk calibration. Our key contributions are as follows:
\begin{itemize}
    \setlength{\itemsep}{2pt}
    \setlength{\parskip}{4pt}
    \item We present a systematic formulation of the conditional risk calibration problem, establishing it as an important question in machine learning.
    \item We analyze conditional risk calibration in both classification and regression settings and demonstrate that the problem can be naturally formulated as a standard regression task. Additionally, for classification settings, we establish a connection between conditional risk calibration and individual probability calibration, showing that improving probability calibration enhances conditional risk estimates.
    \item We provide theoretical and empirical insights into the performance of conditional risk estimators and analyze conditions under which these estimators perform well. Our empirical studies validate these insights and demonstrate their practical relevance in the downstream application of learning to defer (L2D), providing new benchmarks and highlighting the practical benefits of conditional risk calibration for designing more reliable, interpretable, and risk-aware AI systems.
\end{itemize}
We defer the related work section to Appendix \ref{app_relatedwork}.

\section{Formulation}
Let $\mathcal{X}$ denote the input space and $\mathcal{Y}$ the output space. We write $P_{\mathcal{X}}$, $P_{\mathcal{X}\times\mathcal{Y}}$, and $P_{\mathcal{Y}\mid X=x}$ for the marginal, joint, and conditional distributions, respectively. Given a predictor $\hat{f}:\mathcal{X} \to \mathcal{Y}$ and a loss function $\ell:\mathcal{Y}\times\mathcal{Y} \to \mathbb{R}$, we define the conditional loss/risk by
\begin{align}\label{eqn_cond_loss}
    g(x) = \mathbb{E}[\ell(\hat{f}(X), Y)\mid X = x].
\end{align}
With an appropriate $\ell(\cdot)$, $g(x)$ can be interpreted as a measure of the ``uncertainty'' or ``quality'' of the predictor $\hat{f}(\cdot)$. 

In practice, the conditional loss function $g(x)$ is not directly observable, and its estimation is inherently noisy. This challenge is exacerbated in modern AI applications, where inputs often take the form of images or natural language, and for a given input $x$, one typically observes only a single realized loss that provides partial information about $g(x)$.

However, $g(x)$ can be a useful metric in settings like human-AI interactions. For instance, a large $g(x)$ may indicate that $\hat{f}$ is uncertain about the input $x$, and this information can be valuable in several contexts, as discussed below.
 
\paragraph{Uncertainty-based Active Learning.}
Active learning \citep{joshi2009multi,yang2016active,nguyen2022measure} aims to improve model performance while minimizing labeling costs by selectively querying the most informative samples. Conditional risk calibration/estimation provides a principled approach to sample selection by identifying inputs where the predictor $\hat{f}(\cdot)$ exhibits high uncertainty, quantified by the \textit{estimated} conditional loss $\hat{g}(x)$. Data points with high estimated conditional loss $\hat{g}(x)$ are prime candidates for labeling, as they indicate regions where the model is uncertain or prone to error. Acquiring the corresponding label $y$ and retraining on the augmented dataset $(x, y)$ are more likely to improve $\hat{f}$, particularly under a limited labeling budget. By prioritizing these samples, active learning reduces annotation costs while maintaining competitive performance.

\paragraph{Learning to Defer.}
Learning to defer (L2D) \citep{cortes2016learning, madras2018predict} allows an AI system to decide whether to make a prediction or defer to a human expert based on estimated confidence. Conditional risk calibration/estimation provides a natural decision criterion by estimating the expected loss of the model's prediction. If $g(x)$ exceeds a threshold, the system defers to an expert, ensuring higher reliability while minimizing unnecessary interventions. This approach optimally allocates expert resources, reducing deferrals while maintaining high predictive accuracy. It is particularly useful in high-stakes applications.

\subsection{Learning Conditional Loss}\label{subsec_learn_cl}
We now introduce conditional risk calibration/estimation for classification and regression settings.

\paragraph{Classification.} For classification problems, we take $\mathcal{Y} = \{1,2,\cdots, K\}$, where $K$ is the number of classes. Conditional on the input $x$, $Y$ follows a categorical distribution, and the conditional probability is denoted by 
\begin{align}\label{eqn_cond_prob}
    \bm{p}(\cdot|x) = [P(Y=1|X=x),\cdots,P(Y=K|X=x)]
\end{align}
with the shorthand notation $p(x)_{k} = P(Y=k|X=x)$. If we knew $\bm{p}(\cdot\mid x)$, then we could compute the conditional risk as
\begin{equation}\label{eqn:decomp_c}
    \begin{aligned}
    g(x) &= \mathbb{E}[\ell(\hat{f}(X), Y)|X = x] = \sum_{k=1}^K \ell(\hat{f}(x), k)p(x)_{k}.
    \end{aligned}
\end{equation}

\paragraph{Regression.} For regression problems, we have $\mathcal{Y}\subseteq\mathbb{R}$. %We assume $(X,Y)$ follows a general distribution on $P_{\mathcal{X}\times\mathcal{Y}}$, 
We assume that $(X,Y)$ has joint distribution $P_{\mathcal{X}\times\mathcal{Y}}$,
and conditional on $X = x$, we require that the conditional distribution $P_{\mathcal{Y}|x}$ admits a probability density function $p(y|x)$. We then have the following expression for $g(x)$ for regression problems:
\begin{equation}\label{eqn:decomp_r}
    \begin{aligned}
    g(x) &= \mathbb{E}[\ell(\hat{f}(X), Y)|X = x] = \int_{\mathbb{R}} \ell(\hat{f}(x), y)p(y|x)dy.
    \end{aligned}
\end{equation}

\subsubsection{Regression-based Approach for Conditional Risk Estimation}\label{sec_cre_regression}
In this section, we illustrate that the conditional risk estimation problem can be formulated as a standard regression problem in both classification and regression settings. We refer to this as the \textit{regression-based} approach. Under this formulation, the conditional risk calibration/estimation problem enjoys the same statistical guarantees as the standard regression problem.

\paragraph{Conditional Risk as Regression.} Denote by 
$$\mathcal{G} = \{g_{\theta}\mid g_{\theta}: \mathcal{X} \to \mathbb{R}, \theta\in\Theta\}$$ 
the hypothesis class of all functions parametrized by $\theta$ in some parameter space $\Theta$. For both classification and regression settings, given the predictor $\hat{f}$, the dataset $\{x_i, y_i\}_{i=1}^n$ sampled from $P_{\mathcal{X}\times\mathcal{Y}}$, and the loss function $L(\cdot,\cdot): \mathcal{Y}\times\mathcal{Y}\to \mathbb{R}$, the machine learning problem aims to minimize
\begin{align}\label{eqn_ml_goal}
    \min_{\theta\in\Theta} \mathbb{E}_{X\sim P_{\mathcal{X}}} [L(g_{\theta}(X), g(X))].
\end{align}
A standard machine learning approach suggests treating $\{x_i, g(x_i)\}_{i=1}^n$ as the dataset and optimizing the empirical loss $\min_{\theta\in\Theta} \frac{1}{n}\sum_{i=1}^n L(g_{\theta}(x_i), g(x_i))$. 
However, computing $g(x_i)$ is often infeasible, so we must instead rely on its sample estimate $\ell(\hat{f}(x_i), y_i)$, since $\mathbb{E}[\ell(\hat{f}(x_i), y_i)\mid x_i] = g(x_i)$. As a result, one practical approach to estimating the conditional loss is to train on the dataset $\{x_i,z_i\}_{i=1}^n$ with $z_i = \ell(\hat{f}(x_i),y_i)$, optimizing 
\begin{align}\label{eqn_ml}
    \min_{\theta\in\Theta} \frac{1}{n} \sum_{i=1}^n L(g_{\theta}(x_i), z_i),
\end{align}
which is a regression problem. Standard generalization analysis shows that the generalization error of this regression problem depends on the complexity of the hypothesis class $\mathcal{G}$. Denote by
\begin{equation*}
    \begin{aligned}
        R(\theta) &= \mathbb{E}[L(g_{\theta}(X),g(X))] \text{ and }\\
        \hat{R}(\theta) &= \frac{1}{n} \sum_{i=1}^nL(g_{\theta}(x_i),z_i)
    \end{aligned}
\end{equation*}
the population risk and the sample risk, respectively, with $\theta^*$ and $\hat{\theta}$ being the corresponding minimizers. Let 
$$\hat{R}_n(\mathcal{G}) = \mathbb{E}\left[\sup_{\theta\in\Theta}\frac{1}{n}\sum_{i=1}^n\sigma_ig_{\theta}(x_i)\right]$$
denote the empirical Rademacher complexity of $\mathcal{G}$, where $\sigma_i$ are i.i.d. with $P(\sigma_i = 1) = P(\sigma_i = -1) = 1/2$. We present the standard uniform convergence result for the case where $L$ is the $L^1$ loss.
\begin{proposition}\label{prop_generalization}
    Suppose $||\ell(\hat{f}(x), y)||_{\infty} < M$. Then, with probability at least $1-\delta$, for the regression-based approach, the excess risk is bounded by \begin{align}\label{eqn_radmacher_general}
        R(\hat{\theta}) - R(\theta^*) \leq 2\hat{R}_n(\mathcal{G}) + 2M\sqrt{\frac{\ln(1/\delta)}{2n}}.
    \end{align}
\end{proposition}
The proof can be found in Appendix \ref{app_proofs}.

\subsubsection{Calibration-based Approach for Conditional Risk Estimation}\label{sec_cre_calibration}
In this section, we present the \textit{calibration-based} approach for the conditional risk calibration/estimation problem. Note that this approach is \textit{only} applicable to the classification setting. We show that this approach yields improved generalization bounds compared to \eqref{eqn_radmacher_general}, and a more interpretable characterization of the conditional risk.

If we adopt the hypothesis class $$\mathcal{H} = \{\bm{p}_{\theta} | \bm{p}_{\theta} : \mathcal{X} \to [0,1]^K, \theta\in\Theta\}$$
to parametrize the probability mass function $\bm{p}(\cdot|x)$ defined in \eqref{eqn_cond_prob}, then a natural estimator for $g(x)$ is
\begin{align}\label{eqn_gthetax_calibration}
    g_{ca,\theta}(x) = \sum_{k=1}^K \ell(\hat{f}(x), k)p_{\theta}(x)_k,
\end{align}
where the subscript means the calibration-based estimator for $g(\cdot)$. Next, the estimation of the conditional loss depends on how well we can approximate $\bm{p}(x)$ with $\bm{p}_{\theta}(x)$. This is related to the domain of individual calibration \citep{zhao2020individual}, that is we need $||\bm{p}_{\theta}(x) - \bm{p}(x)||$ to have some coverage guarantee for every $x\in\mathcal{X}$.

The following proposition establishes the connection between conditional risk estimation and individual calibration under the \textit{weak realizability} condition.

\begin{definition}
    We say that the distribution $P_{\mathcal{X} \times \mathcal{Y}}$ and the function class $\mathcal{H}$ satisfy weak realizability for the classification problem if, for all $x \in \mathcal{X}$,  
    \begin{equation*}  
    [P(Y = 1 \mid X = x),\cdots,P(Y = K \mid X = x)] \in \mathcal{H}.  
    \end{equation*}  
\end{definition}

An immediate consequence of the weak realizability assumption is that, for classification problems, the (unique) minimizer
\begin{equation} \label{eqn_ml_classification}  
\argmin_{\bm{p}_{\theta}\in\mathcal{H}} \mathbb{E}_{X \sim P_{\mathcal{X}}} [\ell(\bm{p}_{\theta}(X), Y)]  
\end{equation}  
can be used to construct minimizers for the conditional risk \eqref{eqn_ml_goal}. Proposition~\ref{proposition_realizability}, whose proof is given in Appendix~\ref{app_proofs}, formalizes this result.
\begin{proposition}\label{proposition_realizability} 
    Under the weak realizability assumption, and for a suitable loss function $\ell$ such as cross-entropy or the Brier loss, the following holds: 
    \begin{equation*}  
    \sum_{k=1}^K \ell(\hat{f}(\cdot), k) p_{\theta}(\cdot)_k \in \argmin_{g_{\theta} \in \mathcal{G}} \mathbb{E}_{X \sim P_{\mathcal{X}}} [L(g_{\theta}(X), g(X))],  
    \end{equation*}  
    where $\bm{p}_{\theta}(\cdot)$ is the (unique) minimizer of \eqref{eqn_ml_classification}.
\end{proposition}
This proposition implies that in the classification setting, the conditional risk calibration problem can be stated as a calibration problem. Next, we provide theoretical results for the excess risk. Let
$$R_{ca}(\theta) = \mathbb{E}[L(g_{ca,\theta}(X),g(X))]$$
denote the population risk under the calibration-based approach.
\begin{theorem}\label{thm_classification}
For $K$-class classification problems, suppose that $||\ell(\hat{f}(x), y)||_{\infty} < M$. Then, with probability at least $1-\delta$, the excess risk satisfies \begin{align}\label{eqn_radmacher_classification}
        R_{ca}(\hat{\theta}) - R_{ca}(\theta^*) \leq 2M\hat{R}_n(\mathcal{H}) + 2M\sqrt{\frac{\ln(1/\delta)}{2n}}.
    \end{align}
Moreover, for the population loss $R_{ca}(\theta)$, we have for all $\theta \in \Theta$
\begin{equation}\label{eqn_bound_calibration}
        \begin{aligned}
            R_{ca}(\theta) &\leq M \cdot \mathbb{E}[||\bm{p}_{\theta}(X) - \bm{p}(X)||_1]\\
            &\leq M \sqrt{K\left(2 \hat{R}_n(\mathcal{H}) + \sqrt{\frac{\ln(1/\delta)}{2n}}\right)}.
        \end{aligned}
    \end{equation}
\end{theorem}
We refer readers to Appendix \ref{app_proofs} for the proof.

\paragraph{Theoretical Insights.} 
In Theorem \ref{thm_classification}, the bound in \eqref{eqn_bound_calibration} enables a direct control of the population risk $R_{ca}(\theta)$ for all $\theta \in \Theta$, implying that a smaller calibration error leads to a tighter bound on the population risk. This yields stronger theoretical guarantees for the population risk $R_{ca}(\hat{\theta})$, where $\hat{\theta}$ is the empirical risk minimizer. Notably, under a typical classification setup, estimating the conditional probability distribution is often easier than estimating the conditional risk itself. This is because probability distributions tend to exhibit more benign statistical properties, whereas the conditional risk depends intricately on the loss function $\ell(\cdot,\cdot)$, the predictor $\hat{f}$, and the underlying data distribution.

As a result, $R_{ca}(\theta^*)$ is usually lower than $R(\theta^*)$, which implies a lower population risk of $R_{ca}(\hat{\theta})$ compared to  $R(\hat{\theta})$ (see \eqref{eqn_radmacher_general} and \eqref{eqn_radmacher_classification}). We formalize the above statement in the following proposition under a concrete setting. This indicates the theoretical advantages of the calibration-based method over the regression-based method.

\begin{proposition}\label{proposition_population}
Assume the weak realizability condition and suppose that the data distribution is \emph{perfectly separable}; that is, there exists a function $h:\mathcal{X}\to\mathcal{Y}$ such that $\mathbb{P}(h(X)=Y)=1$. 
Let $\ell(\cdot,\cdot)$ be the $0$-$1$ loss. Then, with probability at least $1-\delta$, for every predictor $\hat{f}$, $R_{ca}(\hat{\theta})$ has the upper bound of 
\begin{align*}
    R_{ca}(\hat{\theta}) \leq 2\hat{R}_n(\mathcal{H}) + 2\sqrt{\frac{\ln(1/\delta)}{2n}}.
\end{align*}
In contrast, for regression-based methods, we have
\begin{align*}
    R(\hat{\theta}) \leq R(\theta^*) + 2\hat{R}_n(\mathcal{G}) + 2\sqrt{\frac{\ln(1/\delta)}{2n}},
\end{align*}
and there exists a predictor $\hat{f}$ such that $R(\theta^*) > 0$.
\end{proposition}

In the experimental section, Proposition \ref{proposition_population} is also verified empirically: the calibration-based method consistently achieves lower test empirical risk $\hat{R}(\hat{\theta})$, an unbiased estimator of the population risk $R(\hat{\theta})$, compared to regression-based methods.

\section{Experiments}
In this section, we empirically verify the theoretical insights from the previous section. That is, consistent with Theorem \ref{thm_classification}, we show that
\begin{itemize}
    \item The calibration-based approach can provide better conditional risk estimation than the regression-based approach.
    \item Better calibration for the probability function results in better conditional risk estimation.
\end{itemize}
Moreover, we show that strong conditional risk estimation is beneficial for downstream applications. We conduct extensive experiments on the learning to defer task and obtain results that surpass previous benchmarks.

Regarding the specific experimental goals, we outline the key questions investigated in this section and provide a brief summary of our findings.
\begin{itemize}
    \setlength{\itemsep}{4pt}
    \setlength{\parskip}{4pt}
    \item \textbf{What Improves Conditional Risk Estimation?} Several factors influence conditional risk estimation. First, recall that the conditional loss $g(x)$ in \eqref{eqn_cond_loss} is defined relative to the given predictor $\hat{f}$. A poorly performing predictor increases the scale of $z = \ell(\hat{f}(x), y)$, thereby deteriorating the estimation quality. Second, the choice of loss functions $\ell$ and $L$ also affects the estimation process. Third, the performance of the probability calibrator influences the quality of conditional risk estimation. It is therefore important to disentangle these factors and to understand the performance of different estimation techniques across predictors and loss functions.

    Our empirical results demonstrate that, in the classification setting, calibration-based approaches consistently outperform regression-based methods. Furthermore, improved calibration techniques enhance conditional risk estimation for both weak and strong predictors, showing consistent improvements across different settings.

    \item \textbf{Implications for Downstream Tasks.} Another natural question is \textit{whether better conditional risk estimation leads to better downstream task performance.} To study this, we conduct experiments on learning to defer in both classification and regression settings. Our findings show that the regression-based method is also effective for conditional risk estimation, and that strong conditional risk estimation can enhance performance in learning to defer in both settings. In particular, for the regression with rejection task, our proposed approach outperforms state-of-the-art benchmarks.
\end{itemize}

In Section~\ref{sec_exp_estimation}, we study the classification setting, comparing the calibration-based and regression-based methods for conditional risk estimation. In Section~\ref{sec_exp_l2d}, we study both the classification and regression settings in the downstream application of learning to defer, and experiment with both the calibration-based and regression-based methods.

\subsection{Conditional Risk Estimation for Classification}\label{sec_exp_estimation}
In this section, we investigate the performance of two methods -- calibration-based and regression-based -- for estimating the conditional risk $g(x)$ in the classification setting. 

To show how strong and weak predictors and calibrators (estimators of the conditional risk) affect performance, we conduct experiments using three classifiers:
\begin{itemize}
    \item $\hat{f}_1$, a CNN-based classifier 
    \item $\hat{f}_2$, a ResNet-based classifier 
    \item $\hat{f}_3$, an EfficientNet-based classifier
\end{itemize}
representing low, medium, and high performance tiers, respectively. The goal of our numerical study is to verify the theoretical findings of Propositions~\ref{prop_generalization}-\ref{proposition_realizability} and Theorem~\ref{thm_classification}, which suggest that calibration-based methods outperform regression-based methods for all classifiers $\hat{f}_i$.

\begin{table}[t]
\centering
\caption{Aggregate Results for $\hat{f}_1$ (CNN) Across Different Calibration Approaches. The average loss of $\hat{f}_1$ is 1.30.}
\vspace{0.5em}
\label{tab:cnn_aggregated}
\resizebox{0.95\columnwidth}{!}{
\setlength\extrarowheight{3pt}
\begin{tabular}{c|ccc|ccc}
 \textbf{Loss} & \multicolumn{3}{c|}{\textbf{Calibration-based}} & \multicolumn{3}{c}{\textbf{Regression-based}} \\

 & $\hat{p}_{\theta_1}$  
 & $\hat{p}_{\theta_3}$ 
 & $\hat{p}_{\theta_3, cali}$ 
 & $\hat{g}_{\theta_1}$ 
 & $\hat{g}_{\theta_1, rep}$ 
 & $\hat{g}_{\theta_3}$ \\
\hline
$L^1$ 
 & 0.96   & \textbf{0.28} & 1.71  
 & 2.36  & 1.09  & 2.77  \\
$L^2$ 
 & 4.06  & \textbf{0.97} & 5.13  
 & 10.86 & 4.07  & 15.10 \\
\end{tabular}
}
\end{table}

\paragraph{Calibration-based Methods.}
For a given classifier $\hat{f}$, its conditional loss $g(x)$ can be estimated by replacing the true probabilities $\bm{p}(x)$ in \eqref{eqn:decomp_c} with estimated probabilities $\bm{p}_{\theta}(x)$:
\begin{equation}
\label{eqn:decomp_c_estim}
    g_{\theta}(x) = \sum_{k=1}^K \ell(\hat{f}(x), k)p_{\theta}(x)_k
\end{equation}

Following Proposition~\ref{proposition_realizability},  we can fit probability estimators that minimize the cross-entropy loss or Brier loss. Specifically, to learn $\bm{p}_{\theta}(x)$, we train common computer vision models -- CNN, ResNet, or EfficientNet --
on the training data, and use their softmax outputs as $p_{\theta}(x)$. Thus, we can obtain calibrators $p_{\theta_1}(x)$, $p_{\theta_2}(x)$, and $p_{\theta_3}(x)$,  corresponding to the probability estimates produced by CNN, ResNet, and EfficientNet, respectively. Lastly, the calibration-based estimator is given by the plug-in estimator in \eqref{eqn:decomp_c_estim}.

Furthermore, after obtaining $p_{\theta_i}(x)$, we apply Platt scaling \citep{guo2017calibration}, a common calibration technique, to better reflect the true probability. We denote the resulting temperature-scaled probability estimates by $p_{\theta_i,cal}(x)$.

\paragraph{Regression-based Methods.} 
Instead of using probability estimates, the regression-based method treats calibration of the conditional loss $g(x)$ as a standard regression problem.  Since we focus on a computer vision task in this numerical experiment, we employ suitable architectures to learn $g(x)$. Again, we train three regression models: $g_{\theta_1}(x)$, which is CNN-based; $g_{\theta_2}(x)$, which is ResNet-based; and $g_{\theta_3}(x)$, which is EfficientNet-based. Each model attaches a linear predictor to the last layer of the network and is trained from scratch.

In addition to training a regressor from scratch, we also consider some ``pre-trained'' model, denoted by $g_{\theta, rep}(x)$. We 
first train some strong classifier with cross-entropy loss. Then
$g_{\theta, rep}(x)$ takes the final-layer representation from the classifier as input and performs regression. Since training with cross-entropy loss maximizes the probability of the correct class, we expect that $g_{\theta, rep}(x)$ can benefit from more informed representations compared to $g_{\theta_i}(x)$ trained from scratch. We denote by $g_{\theta_i, rep}(x)$ the regressor that uses the representation learned by $\hat{f}_i$.

\begin{table}[t]
\centering
\caption{Aggregate Results for $\hat{f}_2$ (ResNet) Across Different Calibration Approaches. The average loss of $\hat{f}_2$ is 0.57.}
\vspace{0.5em}
\label{tab:resnet_aggregated}
\resizebox{0.95\columnwidth}{!}{
\setlength\extrarowheight{3pt}
\begin{tabular}{c|ccc|ccc}
 \textbf{Loss} & \multicolumn{3}{c|}{\textbf{Calibration-based}} & \multicolumn{3}{c}{\textbf{Regression-based}} \\

 & $\hat{p}_{\theta_2}$ 
 & $\hat{p}_{\theta_3}$ 
 & $\hat{p}_{\theta_3, cali}$ 
 & $\hat{g}_{\theta_2}$ 
 & $\hat{g}_{\theta_2, rep}$ 
 & $\hat{g}_{\theta_3}$ \\
\hline
$L^1$ 
 & 0.47  & \textbf{0.36} & 1.98  
 & 0.63 & 0.50  & 0.39  \\
$L^2$ 
 & 1.92 & \textbf{1.40} & 5.46  
 & 1.88 & 1.83  & 1.89  \\
\end{tabular}
}
\end{table}

\paragraph{Results.} We conduct experiments for this setting on the CIFAR-10 dataset, where the number of classes $K$ is 10. The results of our numerical study are presented in Tables~\ref{tab:cnn_aggregated}-\ref{tab:resnet_aggregated}. These tables compare the performance of different estimation approaches given various trained classifiers with respect to both $L^1$ and $L^2$ losses.

For calibration-based methods, we observe that better probability estimation -- moving from the lower-performing CNN ($p_{\theta_1}$) to ResNet ($p_{\theta_2}$) and, ultimately, to EfficientNet ($p_{\theta_3}$) -- directly translates into lower calibration errors. This holds for all considered classifiers $\hat{f}$ and for both loss functions. Interestingly, applying Platt scaling worsens calibration performance. This is probably due to the fact that Platt scaling is a group calibration technique rather than an individual calibration technique, leading to distorted probability distributions and, consequently, worse performance in our experiments.

The regression-based method generally performs worse than the calibration-based method. Moreover, its performance in conditional risk estimation is affected by the performance of the predictor $\hat{f}$. In our setting, $\hat{f}_1$ has sample loss $1.3$, whereas $\hat{f}_2$ has sample loss $0.57$, and conditional risk estimation is better for $\hat{f}_2$. These results are again consistent with our theoretical findings. More advanced architectures also lead to reduced loss in conditional risk estimation. Furthermore, using representations from pre-trained classifiers improves performance over training from scratch.

Overall, EfficientNet probability calibration ($p_{\theta_3}$) without Platt scaling achieves the best performance among all considered calibration approaches.

\subsection{Learning to Defer}\label{sec_exp_l2d}

In this section, we begin with the regression setting of the learning to defer (L2D) problem, also known as regression with rejection. We start with this setting because Section~\ref{sec_exp_estimation} already provides sufficient supporting results for the classification setting. After covering regression with rejection, we turn to the classification setting of the L2D problem and present the corresponding results.

\subsubsection{Regression with Rejection}\label{sec_exp_l2d_rwr}

Regression with rejection (RwR) extends the traditional regression framework by introducing a rejector mechanism. Given an input $x$, the system first evaluates $g(x)$, which quantifies the confidence in the regressor's prediction $f(x)$. If the confidence is sufficiently high, the system outputs $f(x)$; otherwise, it defers $x$ to a human expert, who provides a more reliable prediction but at an additional cost $c$.

This approach is particularly valuable in scenarios where incorrect predictions can result in severe losses, and human expertise may be leveraged for uncertain cases.
While most research on learning with rejection has focused on classification problems, there is a growing body of literature exploring regression with rejection 
(\citealp{NIPS2012_ef50c335};
\citealp{pmlr-v97-geifman19a};
\citealp{9207676};
\citealp{NEURIPS2020_e8219d4c};
\citealp{shah2022selective};
\citealp{cheng2024regression};
\citealp{pmlr-v238-li24g}). We now formally define the RwR problem under the fixed-cost setting.

Consider a dataset $\mathcal{D} = \{(x_i, y_i)\}_{i=1}^n$ drawn independently from an unknown distribution $\mathcal{P}_{\mathcal{X}\times\mathcal{Y}}$. The RwR problem involves learning two components: a regressor $f: \mathcal{X} \rightarrow \mathcal{Y}$ and a rejector $r: \mathcal{X} \rightarrow \{0, 1\}$. If $r(x) = 1$, the regressor's prediction for sample $x$ is accepted; if $r(x) = 0$, the sample is deferred to a human expert, incurring a fixed cost of $c>0$. The RwR loss function is designed to balance prediction errors and rejection costs. For example, let $\ell(\cdot, \cdot)$ be the standard $L^2$ loss, given by $\ell(\hat{y}, y) = (\hat{y} - y)^2$. Then the RwR loss function is formulated as:
\begin{equation}
\label{rwr_loss_def}
\ell_{\text{RwR}}(f, r; x, y) = r(x) \cdot \ell(f(x), y) + (1 - r(x)) \cdot c
\end{equation}

Let $\mathcal{G}$ and $\mathcal{R}$ be the sets of candidate regressors and rejectors, respectively. The objective of the RwR problem is to find the optimal regressor-rejector pair $(f, r)$ that minimizes the expected RwR loss~(\ref{rwr_loss_def}):
\begin{equation}\label{eqn_RwR_prob}
\min_{f \in \mathcal{G},\, r \in \mathcal{R}} \, L_{\text{RwR}}(f, r) = \mathbb{E}_{(X, Y) \sim \mathcal{P}_{\mathcal{X}\times\mathcal{Y}}} \left[ \ell_{\text{RwR}}(f, r; X, Y) \right].
\end{equation}

To demonstrate the effectiveness of conditional risk calibration, we focus on a particular RwR setting called \textit{no-rejection learning} \citep{{pmlr-v238-li24g}}, in which the regressor is trained on all available samples as in a standard regression task, while the rejector is learned separately using conditional risk calibration. \textit{No-rejection learning} is a well-studied framework with theoretical guarantees showing that the regressor can achieve optimal performance on the RwR task when complemented by a suitable rejector.
Specifically, suppose that the rejector $r(x)$ for a trained regressor $\hat{f}$ is defined using a trained conditional risk calibrator $g_{\theta}(x)$ as follows:
\begin{equation}
\label{eqn_rejector_def}
    r_{\theta}(x) := \mathbbm{1}\{g_{\theta}(x) \leq c\},
\end{equation}
where the calibrator $g_{\theta}(x)$ is trained to predict the $L^2$ conditional loss of $\hat{f}$. Then the excess RwR loss can be upper-bounded with the following guarantees.
\begin{proposition}[\citealp{pmlr-v238-li24g}] For a given $\hat{f}$ and the calibrator $g_{\theta}(x)$ that predicts the $L^2$ conditional loss of~$\hat{f}$:
\begin{equation*}
\begin{aligned}
&L_{\text{RwR}}(\hat{f}, r_{\theta}) - L_{\text{RwR}}(f^*, r^*) \leq \\ &\quad \mathbb{E}\Bigl[ (\hat{f}(X) - f^*(X))^2 \Bigr] + \mathbb{E}\Bigl[ |g_{\theta}(X) - g(X)| \Bigr],
\end{aligned}
\end{equation*}
where $(f^*, r^*)$ denotes the optimal regressor-rejector pair that minimizes \eqref{eqn_RwR_prob}.
\end{proposition}

These results motivate the application of the conditional risk calibration techniques introduced in Section~\ref{subsec_learn_cl} to the RwR problem \eqref{eqn_RwR_prob} via the no-rejection learning strategy in \eqref{eqn_rejector_def}. In this work, we perform systematic experiments to evaluate different conditional risk calibration strategies under different regressors. 
We demonstrate the benefits of conditional risk calibration for the RwR task by identifying methods that outperform prominent benchmarks across multiple open-source datasets. The remainder of this section conveys the following key messages:
\begin{itemize}
    \item Improved conditional risk calibration leads to better RwR performance for both weak and strong regressors.
    \item Leveraging this insight, we compare our approach with previous RwR algorithms and consistently achieve better performance.  
\end{itemize}

\paragraph{Experimental Setup.}
We conduct experiments on 8 UCI datasets \citep{uci_datasets_ref}, which is a common experiment setup \citep{pmlr-v238-li24g}. %The code used for our study is available at \url{https://github.com/avasi1/conditional-risk-calibration-RwR}. 
For each dataset, we consider 4 fixed rejection costs $c \in \{ 0.2, 0.5, 1.0, 2.0 \}$, resulting in multiple RwR settings defined by the dataset-cost combinations. We then evaluate 16 methods for each RwR setting by training all possible regressor-calibrator pairs $(\hat{f}, g_{\theta})$. We train $\hat{f}$ and $g_{\theta}$ using different models, ranging from simple models with relatively poor performance to more complex models with better predictive accuracy:
\begin{itemize}
    \item LR: Linear regression;
    \item RF: Random forest;
    \item MLP: Multilayer perceptron;
    \item MLP2: Multilayer perceptron with two hidden layers.
\end{itemize}
Since the UCI datasets for the RwR task do not contain images, we do not use ResNet or CNN models. It is generally accepted that for regression tasks, RF-based models and MLP-based models perform relatively well.

All models, except for the random forest, were trained using standardized feature data. We performed 10-fold cross-validation in our experiments. In each fold, the dataset was partitioned into three subsets: 50\% of the data were used to train the regressor $\hat{f}$, 40\% to train the calibrator $g_{\theta}$, and the remaining 10\% were reserved for testing. The full results of our numerical experiments, along with further implementation details,  are provided in Appendix~\ref{app_RwR_results}.

Our systematic study provides insights into the benefits of improved conditional risk calibration and identifies methods that outperform existing benchmarks in the same RwR setting. 

\begin{table}[ht]
\caption{Absolute Errors of Calibrators Given Pretrained Predictors.}
\vspace{0.5em}
\label{tab_calibrators_perf}
\centering
\resizebox{0.95\columnwidth}{!}{
\setlength\extrarowheight{1pt}
\begin{tabular}{c|c|c|cccc}
\multirow{3}{*}{\textbf{Data}} & \multicolumn{1}{c|}{\multirow{3}{*}{\textbf{Predictor}}} & \multicolumn{1}{c|}{\multirow{3}{*}{\makecell{\textbf{Predictor} \\ \textbf{Squared} \\ \textbf{Loss}}}}  & \multicolumn{4}{c}{\textbf{Calibrator Absolute Error}} \\ 
                             &                                    &                                   &             &             &              &             \\
                             &                                    &                                   & \textbf{LR} & \textbf{MLP} & \textbf{MLP2} & \textbf{RF} \\ \hline
\multirow{4}{*}{\rotatebox{90}{\textbf{Concrete}}} & \textbf{LR}   & 112.4 & 107.19 & 96.76  & 99.99 & \textbf{75.37} \\
                                 & \textbf{MLP}  & 83.57 & 80.6   & 73.65  & 77.07 & \textbf{61.01} \\
                                 & \textbf{MLP2} & 58.77 & 59.76  & 55.07  & 56.61 & \textbf{47.94} \\
                                 & \textbf{RF}   & 33.15 & 37.99  & 35.98  & 36.38 & \textbf{32.26} \\ \hline
\multirow{4}{*}{\rotatebox{90}{\textbf{Wine}}}     & \textbf{LR}   & 0.36  & 0.37   & 0.4    & 0.41  & \textbf{0.34} \\
                                 & \textbf{MLP}  & 0.25  & 0.26   & 0.26   & 0.28  & \textbf{0.24} \\
                                 & \textbf{MLP2} & 0.26  & 0.32   & 0.34   & 0.35  & \textbf{0.3}  \\
                                 & \textbf{RF}   & 0.3   & 0.36   & 0.38   & 0.4   & \textbf{0.32} \\ \hline
\multirow{4}{*}{\rotatebox{90}{\textbf{Airfoil}}}  & \textbf{LR}   & 23.55 & 23.49  & 22.72  & 19.2  & \textbf{13.53} \\
                                 & \textbf{MLP}  & 14.01 & 14.32  & 13.41  & 12.25 & \textbf{9.2}  \\
                                 & \textbf{MLP2} & 6.28  & 6.75   & 6.37   & 5.95  & \textbf{5.23} \\
                                 & \textbf{RF}   & 4.62  & 5.29   & 5.08   & 4.81  & \textbf{4.24} \\ \hline
\multirow{4}{*}{\rotatebox{90}{\textbf{Energy}}}   & \textbf{LR}   & 8.24  & 7.87   & 7.65   & 4.28  & \textbf{2.41} \\
                                 & \textbf{MLP}  & 7.54  & 6.74   & 6.53   & 3.67  & \textbf{2.21} \\
                                 & \textbf{MLP2} & 2.36  & 2.01   & 1.91   & 1.64  & \textbf{1.17} \\
                                 & \textbf{RF}   & 0.32  & 0.36   & 0.35   & 0.4   & \textbf{0.31} \\ \hline
\multirow{4}{*}{ \rotatebox{90}{\textbf{Housing}}}  & \textbf{LR}   & 24.96 & 31.73  & 26.28  & 25.53 & \textbf{24.23} \\
                                 & \textbf{MLP}  & 18.06 & 26.89  & \textbf{20.57}  & 22.92 & 20.91 \\
                                 & \textbf{MLP2} & 12.56 & 17.33  & 14.53  & 17.29 & \textbf{14.18} \\
                                 & \textbf{RF}   & 14.56 & 20.44  & 17.47  & 19.04 & \textbf{17.3}  \\ \hline
\multirow{4}{*}{\rotatebox{90}{\textbf{Solar}}}    & \textbf{LR}   & 0.63  & \textbf{0.81}   & 0.91   & 0.96  & 0.85 \\
                                 & \textbf{MLP}  & 0.66  & \textbf{0.86}   & 0.94   & 1     & 0.88 \\
                                 & \textbf{MLP2} & 0.7   & \textbf{0.92}   & 0.99   & 1.04  & 0.94 \\
                                 & \textbf{RF}   & 0.81  & \textbf{1.05}   & 1.12   & 1.21  & 1.07 \\ \hline
\multirow{4}{*}{\rotatebox{90}{\textbf{Forest}}}   & \textbf{LR}   & 2.11  & \textbf{2.02}   & 2.19   & 2.76  & 2.22 \\
                                 & \textbf{MLP}  & 2.09  & \textbf{2.02}   & 2.12   & 2.86  & 2.18 \\
                                 & \textbf{MLP2} & 2.12  & \textbf{2.03}   & 2.15   & 2.96  & 2.19 \\
                                 & \textbf{RF}   & 2.22  & \textbf{2.23}   & 2.33   & 3.06  & 2.35 \\ \hline
\multirow{4}{*}{\rotatebox{90}{\textbf{Parkinsons }}} & \textbf{LR}   & 86.53 & 76.24  & 66.75  & 37.07 & \textbf{21.23} \\
                                 & \textbf{MLP}  & 19.87 & 21.4   & 19.43  & 18.68 & \textbf{16.17} \\
                                 & \textbf{MLP2} & 6.62  & 8.82   & 7.82   & 8.54  & \textbf{7.07}  \\
                                 & \textbf{RF}   & 0.35  & 0.57   & 0.75   & 0.74  & \textbf{0.45}  \\
\end{tabular}
}
\label{tab:predictor-calibrator}
\end{table}

\paragraph{Model Performance for Predictors and Calibrators.}\label{exp_calibration}
To facilitate interpretation of the results, we adopt a naming scheme in which each model is denoted by its regressor–calibrator combination. For example, MLP+RF indicates that the regressor $\hat{f}$ is a multilayer perceptron with one hidden layer, and the calibrator $g_{\theta}$ is a random forest.

We first identify strong and weak predictors and calibrators among our models. Table \ref{tab:predictor-calibrator} indicates that LR generally performs relatively poorly as both a predictor and a calibrator, whereas RF and MLP2 exhibit comparatively better performance in both roles. The only exceptions are the {\scshape Solar} and {\scshape Forest} datasets, where LR demonstrates strong performance due to their target distributions, with a single target value dominating the data -- especially in the case of {\scshape Solar}, where 83\% of targets correspond to just one value. In cases where all models yield similar estimations (e.g., {\scshape Wine}, {\scshape Solar}, and {\scshape Forest}), the predictor loss and calibrator loss are relatively small in scale compared to those for other datasets. This analysis enables a general categorization of the four models based on their effectiveness as predictors and calibrators.

\begin{table}[ht]
\caption{Relationship Between Calibrator Performance and RwR Loss for the {\scshape Energy} Dataset.}
\label{calibrator_vs_RwR_table_energy}
\vspace{0.5em}
\centering
\setlength\extrarowheight{2pt}
\resizebox{1.0\columnwidth}{!}{
\begin{tabular}{c|c||llll}
 & \multicolumn{1}{c||}{} & \multicolumn{4}{c}{\textbf{RwR Loss}} \\
\multirow{-2}{*}{\textbf{}} & \multicolumn{1}{c||}{\multirow{-2}{*}{\makecell{\textbf{Calibrator}\\ \textbf{Absolute Error}}}} & \multicolumn{1}{c}{$c=0.2$} & \multicolumn{1}{c}{$c=0.5$} & \multicolumn{1}{c}{$c=1.0$} & \multicolumn{1}{c}{$c=2.0$} \\ \hline
\textbf{LR+LR} & \cellcolor[HTML]{F8696B}\textbf{7.87} & \cellcolor[HTML]{F8696B}0.77 & \cellcolor[HTML]{F8696B}1.03 & \cellcolor[HTML]{F8696B}1.46 & \cellcolor[HTML]{F8696B}2.17 \\
\textbf{LR+MLP} & \cellcolor[HTML]{F97A7D}\textbf{7.65} & \cellcolor[HTML]{FBD1D4}0.52 & \cellcolor[HTML]{FBD4D7}0.79 & \cellcolor[HTML]{FCD8DB}1.20 & \cellcolor[HTML]{FBC3C5}1.93 \\
\textbf{LR+MLP2} & \cellcolor[HTML]{AFC5E3}\textbf{4.28} & \cellcolor[HTML]{8EAED8}0.31 & \cellcolor[HTML]{A0BBDE}0.61 & \cellcolor[HTML]{C8D7EC}1.03 & \cellcolor[HTML]{C8D7EC}1.62 \\
\textbf{LR+RF} & \cellcolor[HTML]{5A8AC6}\textbf{2.41} & \cellcolor[HTML]{5A8AC6}0.26 & \cellcolor[HTML]{5A8AC6}0.54 & \cellcolor[HTML]{5A8AC6}0.85 & \cellcolor[HTML]{5A8AC6}1.29 \\ \hline
\textbf{MLP+LR} & \cellcolor[HTML]{F8696B}\textbf{6.74} & \cellcolor[HTML]{F8696B}0.44 & \cellcolor[HTML]{F8696B}0.69 & \cellcolor[HTML]{FBD8DA}1.06 & \cellcolor[HTML]{FCF2F5}1.65 \\
\textbf{MLP+MLP} & \cellcolor[HTML]{F97C7E}\textbf{6.53} & \cellcolor[HTML]{F97A7C}0.43 & \cellcolor[HTML]{F8696B}0.69 & \cellcolor[HTML]{F8696B}1.18 & \cellcolor[HTML]{F8696B}1.97 \\
\textbf{MLP+MLP2} & \cellcolor[HTML]{ABC3E2}\textbf{3.67} & \cellcolor[HTML]{A5BFE0}0.27 & \cellcolor[HTML]{BCCFE8}0.56 & \cellcolor[HTML]{DEE7F4}0.98 & \cellcolor[HTML]{EEF2FA}1.60 \\
\textbf{MLP+RF} & \cellcolor[HTML]{5A8AC6}\textbf{2.21} & \cellcolor[HTML]{5A8AC6}0.20 & \cellcolor[HTML]{5A8AC6}0.46 & \cellcolor[HTML]{5A8AC6}0.80 & \cellcolor[HTML]{5A8AC6}1.32 \\ \hline
\textbf{MLP2+LR} & \cellcolor[HTML]{F8696B}\textbf{2.01} & \cellcolor[HTML]{FCFCFF}0.23 & \cellcolor[HTML]{FCFCFF}0.49 & \cellcolor[HTML]{F8696B}0.79 & \cellcolor[HTML]{F8696B}1.18 \\
\textbf{MLP2+MLP} & \cellcolor[HTML]{FAA8AA}\textbf{1.91} & \cellcolor[HTML]{FCFCFF}0.23 & \cellcolor[HTML]{FCFCFF}0.49 & \cellcolor[HTML]{EFF3FA}0.77 & \cellcolor[HTML]{FBD8DA}1.15 \\
\textbf{MLP2+MLP2} & \cellcolor[HTML]{D7E2F2}\textbf{1.64} & \cellcolor[HTML]{F8696B}0.27 & \cellcolor[HTML]{F8696B}0.50 & \cellcolor[HTML]{FBCCCE}0.78 & \cellcolor[HTML]{EBF0F9}1.13 \\
\textbf{MLP2+RF} & \cellcolor[HTML]{5A8AC6}\textbf{1.17} & \cellcolor[HTML]{5A8AC6}0.20 & \cellcolor[HTML]{5A8AC6}0.45 & \cellcolor[HTML]{5A8AC6}0.71 & \cellcolor[HTML]{5A8AC6}1.04 \\ \hline
\textbf{RF+LR} & \cellcolor[HTML]{FCECEF}\textbf{0.36} & \cellcolor[HTML]{F8696B}0.17 & \cellcolor[HTML]{F8696B}0.28 & \cellcolor[HTML]{C6D6EC}0.31 & \cellcolor[HTML]{FBCCCE}0.33 \\
\textbf{RF+MLP} & \cellcolor[HTML]{E9EFF8}\textbf{0.35} & \cellcolor[HTML]{DBE5F3}0.16 & \cellcolor[HTML]{C6D6EC}0.27 & \cellcolor[HTML]{F8696B}0.32 & \cellcolor[HTML]{5A8AC6}0.32 \\
\textbf{RF+MLP2} & \cellcolor[HTML]{F8696B}\textbf{0.40} & \cellcolor[HTML]{F8696B}0.17 & \cellcolor[HTML]{F8696B}0.28 & \cellcolor[HTML]{F8696B}0.32 & \cellcolor[HTML]{F8696B}0.34 \\
\textbf{RF+RF} & \cellcolor[HTML]{5A8AC6}\textbf{0.31} & \cellcolor[HTML]{5A8AC6}0.14 & \cellcolor[HTML]{5A8AC6}0.26 & \cellcolor[HTML]{5A8AC6}0.30 & \cellcolor[HTML]{5A8AC6}0.32
\end{tabular}
}
\end{table}

\paragraph{Relationship Between Calibration Quality and RwR Performance.}\label{exp_calibration_RwR}

\begin{figure*}[t]
\centering
\includegraphics[width=0.9\textwidth]{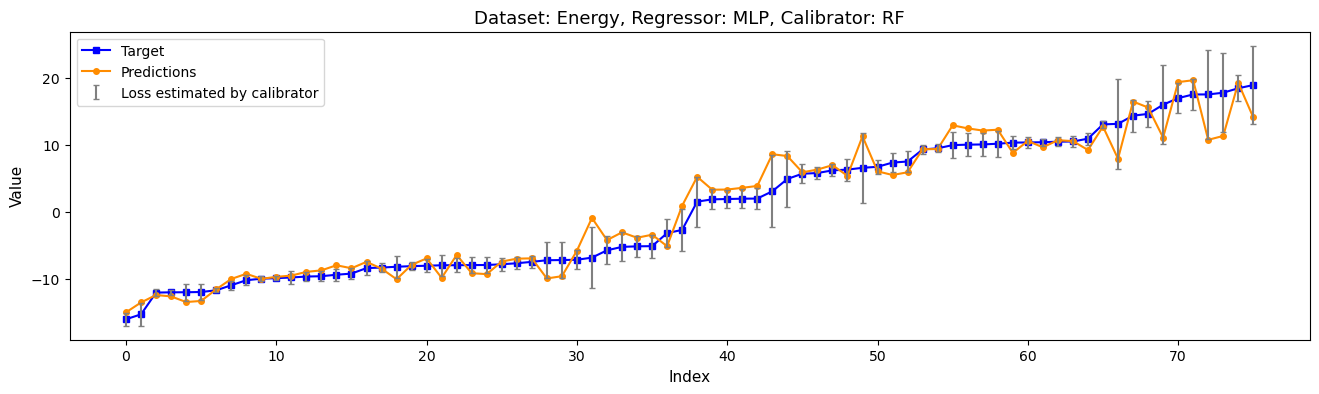}
    \caption{MLP Losses Estimated by an RF Calibrator on the {\scshape Energy} Test Data.}
\label{energy_calibrator_illustration}
\end{figure*}

Our study also shows that, within a group of calibration methods under a fixed RwR setting, better calibration leads to improved RwR performance. Table~\ref{calibrator_vs_RwR_table_energy} presents the results for the {\scshape Energy} dataset across different rejection costs. In particular, for a fixed RwR setting and regressor, we compute the mean absolute errors and the RwR losses that correspond to each calibrator, and examine the relationship between these two metrics. To facilitate comparison, we highlight lower values in blue and higher values in red within each group of calibrators (sharing the same RwR setting and regressor), making it easy to observe the relationship between calibrators' errors and the RwR losses. This example clearly demonstrates that lower calibrator errors generally correspond to lower RwR costs.
This finding further supports the use of conditional risk calibration in downstream tasks and corroborates the theoretical findings of \cite{pmlr-v238-li24g}.

More specifically, the RF calibrator outperforms other calibrators on the vast majority of the considered datasets (see Table~\ref{tab_calibrators_perf}). To further illustrate its performance, we present Figure~\ref{energy_calibrator_illustration}, which shows the results obtained using the RF calibrator with a trained MLP regressor on the {\scshape Energy} test data (from one fold). The plot displays the true target values $f(x)$, the predictions made by the MLP regressor $\hat{f}(x)$, and the square root of the losses estimated by the RF calibrator $g_{\theta}(x)$. The test instances are sorted in ascending order of the target value for clearer visualization. By comparing the calibrator's estimates with the sample conditional risk, we observe that the calibrator performs exceptionally well, accurately reflecting the MLP prediction errors.

\begin{table*}[ht]
\caption{Comparative Performance of Learning with Rejection Methods, Measured by RwR Losses on the Testing Data.}
\label{RwR_selection_table_horizontal}
\centering
\resizebox{0.9\textwidth}{!}{
\setlength\extrarowheight{3pt}
\begin{tabular}{l|llll|llll|llll|llll}
\textbf{} & \multicolumn{4}{c|}{\scshape{Concrete}} & \multicolumn{4}{c|}{\scshape{Wine}} & \multicolumn{4}{c|}{\scshape{Airfoil}} & \multicolumn{4}{c}{\scshape{Energy}} \\
\textbf{Method\textbackslash{}Cost $c$} & 0.2 & 0.5 & 1.0 & 2.0 & 0.2 & 0.5 & 1.0 & 2.0 & 0.2 & 0.5 & 1.0 & 2.0 & 0.2 & 0.5 & 1.0 & 2.0 \\ \hline \rowcolor{gray!20}
\textbf{MLP+RF} & \textbf{0.20} & \textbf{0.50} & \textbf{1.00} & \textbf{2.00} & \textbf{0.17} & \textbf{0.24} & \textbf{0.25} & \textbf{0.25} & \textbf{0.20} & 0.50 & 0.99 & 1.91 & 0.20 & 0.46 & 0.80 & 1.32 \\ \rowcolor{gray!20}
\textbf{RF+RF} & \textbf{0.20} & \textbf{0.50} & \textbf{1.00} & \textbf{2.00} & 0.18 & 0.28 & 0.30 & 0.31 & \textbf{0.20} & 0.50 & \textbf{0.97} & \textbf{1.73} & \textbf{0.14} & \textbf{0.26} & \textbf{0.30} & \textbf{0.32} \\
\textbf{NN+kNNRej} & \textbf{0.20} & \textbf{0.50} & \textbf{1.00} & \textbf{2.00} & 0.23 & 0.27 & 0.28 & 0.28 & \textbf{0.20} & 0.50 & 1.00 & 2.02 & 0.21 & 0.58 & 1.03 & 1.60 \\
\textbf{SelNet} & 0.75 & 1.09 & 2.82 & 3.16 & 0.18 & \textbf{0.24} & \textbf{0.25} & \textbf{0.25} & 0.21 & \textbf{0.49} & 0.99 & 1.84 & 0.26 & 0.49 & 0.86 & 1.39 \\ \hline
\end{tabular}
}

\resizebox{0.9\textwidth}{!}{
\setlength\extrarowheight{3pt}
\begin{tabular}{l|llll|llll|llll|llll}
\textbf{} & \multicolumn{4}{c|}{\scshape{Housing}} & \multicolumn{4}{c|}{\scshape{Solar}} & \multicolumn{4}{c|}{\scshape{Forest}} & \multicolumn{4}{c}{\scshape{Parkinsons}} \\
\textbf{Method\textbackslash{}Cost $c$} & 0.2 & 0.5 & 1.0 & 2.0 & 0.2 & 0.5 & 1.0 & 2.0 & 0.2 & 0.5 & 1.0 & 2.0 & 0.2 & 0.5 & 1.0 & 2.0 \\ \hline \rowcolor{gray!20}
\textbf{MLP+RF} & \textbf{0.20} & \textbf{0.50} & \textbf{1.00} & \textbf{2.00} & 0.23 & \textbf{0.40} & 0.54 & \textbf{0.64} & \textbf{0.20} & 0.51 & 1.08 & \textbf{1.97} & 0.20 & 0.50 & 1.00 & 2.00 \\ \rowcolor{gray!20}
\textbf{RF+RF} & \textbf{0.20} & \textbf{0.50} & 1.01 & 2.07 & 0.23 & 0.41 & 0.57 & 0.73 & \textbf{0.20} & 0.51 & \textbf{1.04} & 1.98 & \textbf{0.12} & \textbf{0.20} & \textbf{0.26} & \textbf{0.32} \\
\textbf{NN+kNNRej} & 0.22 & 0.59 & \textbf{1.00} & \textbf{2.00} & 0.19 & \textbf{0.40} & 0.60 & 0.75 & 0.30 & \textbf{0.50} & 1.09 & 2.13 & 0.20 & 0.50 & 1.00 & 2.00 \\
\textbf{SelNet} & 1.27 & 1.31 & 2.23 & 2.47 & \textbf{0.18} & 0.45 & \textbf{0.52} & 0.68 & 0.54 & 0.78 & 1.25 & 2.17 & 0.20 & 0.60 & 1.17 & 1.97 \\
\end{tabular}
}
\end{table*}

\paragraph{Advantage Over Existing Results.}\label{exp_calibration_advantage}
According to Table~\ref{calibrator_vs_RwR_table_energy}, the two methods that exhibit the best overall performance in our analysis are MLP+RF and RF+RF. In Table~\ref{RwR_selection_table_horizontal}, we compare the performance of these methods against the following two RwR benchmarks with respect to the RwR loss~(\ref{rwr_loss_def}): SelectiveNet \citep{pmlr-v97-geifman19a} and NN+kNNRej \citep{pmlr-v238-li24g}. SelectiveNet is a deep learning approach with an integrated rejector, whereas NN+kNNRej is a \textit{no-rejection learning} algorithm that achieved the best performance in the study of \cite{pmlr-v238-li24g}.

Table~\ref{RwR_selection_table_horizontal} shows that MLP+RF matches or surpasses NN+kNNRej in 30 out of 32 RwR settings, and in 27 out of 32 settings when compared to SelectiveNet. In the remaining cases, the differences in RwR losses are generally small. Similarly, RF+RF matches or surpasses NN+kNNRej in 24 out of 32 settings, and in 25 out of 32 settings when compared to SelectiveNet. It is worth noting that for the {\scshape Parkinsons} dataset, the RF+RF method substantially outperforms the other methods, which can be explained by the superior performance of the regressor.

\paragraph{Conclusion.} The experiments in Section~\ref{sec_exp_l2d_rwr} show that better conditional risk calibration can improve the RwR loss for both strong and weak predictors. By adopting a stronger conditional risk calibrator, our method consistently outperforms existing benchmarks on the RwR task.

\subsubsection{Classification Setting for L2D}\label{sec_exp_l2d_cla}
Lastly, we conduct the L2D experiment for the classification setting.
We use the standard L2D loss definition as in \eqref{rwr_loss_def} and \eqref{eqn_rejector_def}, with the predictor now being a classification model.

\paragraph{Experimental Setup.} We employ two predictors with varying performance: a CNN model as the weak predictor and a ResNet model as the strong predictor. For the calibrator, similarly to Section~\ref{sec_exp_estimation}, we adopt both calibration-based and regression-based methods. For each predictor, we train three calibration-based methods, where the estimated probabilities are taken from the outputs of the CNN, ResNet, and EfficientNet models, respectively. For the regression-based approach, based on the results in Tables~\ref{tab:cnn_aggregated} and \ref{tab:resnet_aggregated}, we extract representations from an EfficientNet classifier and use them as input to a regressor for conditional risk estimation.

We present the calibrator’s absolute error (relative to the true loss) and the corresponding L2D loss in Table~\ref{calibrator_vs_RwR_table_cifar10} for various predictor–calibrator combinations. Similar to the findings in previous sections, for both CNN and ResNet predictors, using better calibration models leads to a reduction in L2D loss. Lastly, the representation-based regression approach performs poorly in terms of both calibrator absolute error and L2D loss.

\begin{table}[t]
\caption{Relationship Between Calibrator Performance and L2D Loss for the CIFAR-10 Dataset. Here, EFFNET stands for EfficientNet.}
\label{calibrator_vs_RwR_table_cifar10}
\vspace{\baselineskip}
\centering
\setlength\extrarowheight{2pt}
\resizebox{1.0\columnwidth}{!}{
\begin{tabular}{c|c||llll}
 & \multicolumn{1}{c||}{} & \multicolumn{4}{c}{\textbf{L2D Loss}} \\
\multirow{-2}{*}{\textbf{}} & \multicolumn{1}{c||}{\multirow{-2}{*}{\makecell{\textbf{Calibrator}\\ \textbf{Absolute Error}}}} & \multicolumn{1}{c}{$c=0.2$} & \multicolumn{1}{c}{$c=0.5$} & \multicolumn{1}{c}{$c=1.0$} & \multicolumn{1}{c}{$c=2.0$} \\ \hline

\textbf{CNN+CNN}       & \textbf{0.76} & 0.20 & 0.41 & 0.65 & 0.75 \\
\textbf{CNN+RESNET}      & \textbf{0.45} & \bf{0.15} & \bf{0.32} & \bf{0.51} & \bf{0.70} \\
\textbf{CNN+EFFNET}     & \textbf{0.47} & 0.16 & \bf{0.32} & 0.53 & 0.74 \\
\textbf{CNN+FEATURE}       & \textbf{1.49} & 0.28 & 0.47 & 0.67 & 0.82 \\ \hline

\textbf{RESNET+CNN}      & \textbf{1.48} & 0.17 & 0.36 & 0.60 & 0.88 \\
\textbf{RESNET+RESNET}     & \textbf{0.48} & 0.22 & 0.38 & 0.53 & \bf{0.57} \\
\textbf{RESNET+EFFNET}    & \textbf{0.62} & \bf{0.13} & \bf{0.26} & \bf{0.42} & 0.61 \\
\textbf{RESNET+FEATURE}      & \textbf{1.03} & 0.25 & 0.40 & 0.56 & 0.68 \vspace{1em}\\

\end{tabular}
}
\end{table}

\section{Conclusion}
In this work, we formulated and analyzed the problem of conditional risk calibration, establishing its theoretical foundations and practical significance. Our theoretical analysis is supported by empirical studies that validate our insights and demonstrate the effectiveness of improved conditional risk estimation in downstream applications. By bridging theoretical understanding and real-world applications, our work lays the foundation for future advances in uncertainty-aware decision-making and for the development of more reliable, interpretable, and risk-aware machine learning systems.

\bibliography{references_calibrating_conditional_risk.bib}

\clearpage
\appendix
\onecolumn

\begin{center}
{\LARGE Calibrating Conditional Risk\par}
\vspace{0.5em}
{\Large Supplementary Materials\par}
\end{center}

\section{Related Work}\label{app_relatedwork}

\paragraph{Uncertainty Estimation.}
While the conditional risk estimation/calibration problem has not been explicitly studied in prior work, it is closely related to established areas such as uncertainty quantification and calibration. One of the most common approaches to uncertainty quantification is probability calibration, where a model's predicted probabilities are adjusted to reflect true likelihoods. Methods such as Platt scaling \citep{platt1999probabilistic}, isotonic regression \citep{zadrozny2002transforming}, and temperature scaling \citep{guo2017calibration} have been widely used to improve the reliability of predicted probabilities. However, these methods focus on calibrating probabilities rather than directly estimating the expected loss, making them insufficient for conditional risk estimation. Recent work on conformal prediction \citep{angelopoulos2023conformal} and individual calibration \citep{zhao2020individual} has explored methods for instance-level uncertainty estimation, but these approaches typically rely on confidence scores rather than on explicit loss estimation.

\paragraph{Human-AI Interaction.}
Our study is related to the Learning to Defer (L2D) setting \citep{madras2018predict,verma2022calibrated,mozannar2023should}, which belongs to the broader field of Human-AI Interaction. Similar problems have been studied under related frameworks, including learning under triage \citep{okati2021differentiable}, learning under human assistance \citep{de2020regression}, learning to complement humans \citep{wilder2020learning}, and selective prediction \citep{pmlr-v97-geifman19a}, all of which share a similar setup.

\paragraph{Learning to Defer.}
We conduct experiments in the setting where the cost of deferral is constant, which has a long history in machine learning and is known as rejection learning \citep{chow1970optimum,bartlett2008classification,cortes2016learning}. More recent developments focus on various aspects, including the one-stage vs.\ two-stage process \citep{raghu2019algorithmic,mao2024predictor}, mixture of experts \citep{pradier2021preferential,wilder2020learning}, and the related surrogate loss \citep{charusaie2022sample,verma2022calibrated,keswani2021towards,mao2024theoretically}.

\paragraph{Regression with Rejection.}
Regression with Rejection is a special setting of L2D. In this setting, \cite{NEURIPS2020_e8219d4c} characterize the optimal predictor and rejector for the class of all measurable functions, and propose a nonparametric algorithm to learn them. \cite{kang2023surrogate} quantify the prediction uncertainty directly, which can also be applied to address the regression with rejection problem. \cite{jiang2020risk} aim to minimize the rejection rate under a specified level of loss without considering rejection costs. \cite{shah2022selective} consider both rejection costs and reject budgets, highlighting the challenges of optimizing the predictor and rejector even for the training samples, and develop a greedy algorithm to solve the problem approximately. For additional discussion of the regression with rejection problem, see also \cite{pmlr-v238-li24g,cheng2024regression}.

\section{Proofs}\label{app_proofs}
\subsection{Proof of Proposition \ref{prop_generalization}}

Using a standard telescoping sum, we have
\begin{align*}
    R(\hat{\theta}) - R(\theta^*)
    &= R(\hat{\theta}) - \hat{R}(\hat{\theta})
    + \hat{R}(\hat{\theta}) - \hat{R}(\theta^*)
    + \hat{R}(\theta^*) - R(\theta^*) \\
    &\leq R(\hat{\theta}) - \hat{R}(\hat{\theta})
    + \hat{R}(\theta^*) - R(\theta^*) \\
    &\leq 2 \sup_{\theta \in \Theta} |R(\theta) - \hat{R}(\theta)|.
\end{align*}
Thus, the excess risk is reduced to bounding the uniform deviation term, which is a standard approach in statistical learning theory. By the standard symmetrization argument \citep{mohri2018foundations}, it follows that
$$
\mathbb{E}\left[\sup_{\theta \in \Theta} |R(\theta) - \hat{R}(\theta)|\right]
\leq 2 \mathbb{E}[\hat{R}_n(\mathcal{L})],
$$
where $\mathcal{L}$ is the loss function class
$$
\mathcal{L}
=
\left\{
L(\cdot,\cdot) : \mathcal{X}\times\mathcal{Y}  \to \mathbb{R}
\;\middle|\;
L(x,z) = |g_{\theta}(x) - z|,\ \text{where } g_{\theta} \in \mathcal{G}
\right\}.
$$
Since the function $u \mapsto |u-z|$ is $1$-Lipschitz for every fixed $z$, the contraction inequality gives
$$
\mathbb{E}[\hat{R}_n(\mathcal{L})]
\leq 1\cdot \mathbb{E}[\hat{R}_n(\mathcal{G})].
$$
Hence,
$$
\mathbb{E}\left[\sup_{\theta \in \Theta} |R(\theta) - \hat{R}(\theta)|\right]
\leq 2 \mathbb{E}[\hat{R}_n(\mathcal{G})].
$$

Recall that $\hat{R}_n(\mathcal{G})$ is the empirical Rademacher complexity,
$$
\hat{R}_n(\mathcal{G})
=
\mathbb{E}_{\sigma}\left[
\sup_{\theta \in \Theta}
\frac{1}{n}\sum_{i=1}^n \sigma_i g_{\theta}(x_i)
\right],
$$
where $\sigma_i$ are i.i.d. Rademacher random variables satisfying
$P(\sigma_i = 1) = P(\sigma_i = -1) = 1/2$.

Since $||\ell(\hat{f}(x), y)||_{\infty} < M$, we have $\ell(\hat{f}(x_i), y_i) \leq M$. It then follows from a standard bounded-difference inequality that with probability at least $1-\delta$,
$$
\sup_{\theta \in \Theta} |R(\theta) - \hat{R}(\theta)|
\leq
2\hat{R}_n(\mathcal{G})
+
M\sqrt{\frac{\ln(1/\delta)}{2n}}.
$$

\subsection{Proof of Proposition \ref{proposition_realizability}}

By the weak realizability assumption, the conditional probability vector $\bm{p}(x)$ belongs to the class $\mathcal{H}$. Because $\ell$ is assumed to be either the cross-entropy loss or the Brier score, both of which are proper classification losses, it follows that $\bm{p}(x)$ uniquely minimizes the pointwise expected loss. Thus,
$$
\bm{p}(x) \in \argmin_{\bm{p}_{\theta}(x) \in \mathcal{H}}\mathbb{E}_{X \sim P_{\mathcal{X}}} [\ell(\bm{p}_{\theta}(X), Y)].
$$
Consequently, we denote the minimizer by $\bm{p}_{\theta^*}(x) = \bm{p}(x)$. The function $\bm{p}_{\theta^*}$ induces
$$
g_{\theta^*}(x) := \sum_{k=1}^K \ell(\hat{f}(x), k)p_{\theta^*}(x)_k.
$$
Notice that $g_{\theta^*}(x)$ exactly coincides with the true conditional loss $g(x)$, since
$$
g(x) = \sum_{k=1}^K \ell(\hat{f}(x), k)p(x)_k.
$$
Therefore, we have
$L(g_{\theta^*}(x), g(x)) = 0$
for all $x \in \mathcal{X}$. Lastly, it follows that
$\mathbb{E}_{X \sim P_{\mathcal{X}}}[L(g_{\theta^*}(X), g(X))] = 0,$
reaching the minimum for any standard loss function $L$. Therefore,
$$
\sum_{k=1}^K \ell(\hat{f}(\cdot), k) p_{\theta^*}(\cdot)_k = g_{\theta^*}\in \argmin_{g_{\theta} \in \mathcal{G}} \mathbb{E}_{X \sim P_{\mathcal{X}}} [L(g_{\theta}(X), g(X))].
$$

\subsection{Proof of Theorem \ref{thm_classification}}

The proof of \eqref{eqn_radmacher_classification} is the same as that of Proposition \ref{prop_generalization}. One only needs to observe that in the classification setting, we have
$$
g_{ca,\theta}(x)
\leq \sum_{k=1}^K |\ell(\hat{f}(x), k)| p_{\theta}(x)_k
\leq M \sum_{k=1}^K p_{\theta}(x)_k
= M.
$$

For \eqref{eqn_bound_calibration}, note that the first inequality follows from
$$
|g_{ca,\theta}(x) - g(x)|
=
\left|\sum_{k=1}^K \ell(\hat{f}(x), k)\bigl(p_{\theta}(x)_k - p(x)_k\bigr)\right|
\leq
M \sum_{k=1}^K \left|p_{\theta}(x)_k - p(x)_k\right|
=
M\|\bm{p}_{\theta}(x) - \bm{p}(x)\|_1.
$$

For the second inequality, it suffices to bound $\|\bm{p}_{\theta}(x) - \bm{p}(x)\|_1$ by a suitable excess risk, and then apply the uniform convergence bound.

First, to relate this quantity to the risk, we need to work on the $L^2$ distance instead of $L^1$. Using the bound
$\|\bm{p}_{\theta}(x) - \bm{p}(x)\|_1
\leq
\sqrt{K}\,\|\bm{p}_{\theta}(x) - \bm{p}(x)\|_2$
together with Jensen's inequality
$\mathbb{E}[\|\bm{v}\|_2] \leq \sqrt{\mathbb{E}[\|\bm{v}\|_2^2]},$
we obtain
$$
\mathbb{E}[\|\bm{p}_{\theta}(X) - \bm{p}(X)\|_1]
\leq
\sqrt{K\,\mathbb{E}[\|\bm{p}_{\theta}(X) - \bm{p}(X)\|_2^2]}.
$$

Next, to incorporate the population risk, consider the loss function
$\ell(\bm{p}_{\theta}(x), \bm{y}) = \|\bm{p}_{\theta}(x) - \bm{y}\|_2^2,$
where $\bm{y}$ is the one-hot vector corresponding to the label of $x$. We then define
$$
R(\theta) = \mathbb{E}[\|\bm{p}_{\theta}(x) - \bm{y}\|_2^2].
$$
Since the weak realizability condition is assumed throughout, we adopt the convention that there exists $\theta^* \in \Theta$ such that $\bm{p}_{\theta^*}(x) = \bm{p}(x)$, and denote
$R(\theta^*) = \mathbb{E}[\|\bm{p}(x) - \bm{y}\|_2^2].$

Then, by the geometric property of the $L^2$ distance,
$$
\|\bm{p}_{\theta}(x) - \bm{y}\|_2^2
=
\|\bm{p}_{\theta}(x) - \bm{p}(x)\|_2^2
+
\|\bm{p}(x) - \bm{y}\|_2^2
+
(\bm{p}_{\theta}(x) - \bm{p}(x))^\top (\bm{p}(x) - \bm{y}).
$$
Using the fact that $\mathbb{E}[\bm{p}(x) - \bm{y}] = 0$, we obtain
$$
\|\bm{p}_{\theta}(x) - \bm{p}(x)\|_2^2
=
\|\bm{p}_{\theta}(x) - \bm{y}\|_2^2
-
\|\bm{p}(x) - \bm{y}\|_2^2.
$$
This establishes the identity
\begin{align}\label{eqn:brier}
    \mathbb{E}[\|\bm{p}_{\theta}(X) - \bm{p}(X)\|_2^2] = R(\theta) - R(\theta^*).
\end{align}

Therefore,
$$
R_{ca}(\theta)
\leq
M\,\mathbb{E}[\|\bm{p}_{\theta}(X) - \bm{p}(X)\|_1]
\leq
M \sqrt{K(R(\theta) - R(\theta^*))}.
$$
Finally, the excess risk $R(\theta) - R(\theta^*)$ can be bounded by the same uniform convergence argument as in Proposition~\ref{prop_generalization}, and the estimation target is a probability function bounded by $1$. Repeating the same argument yields the bound of 
$2 \hat{R}_n(\mathcal{H}) + \sqrt{\frac{\ln(1/\delta)}{2n}}$.

\subsection{Proof of Proposition \ref{proposition_population}}

For the calibration-based method, under the weak realizability condition and the perfect separability condition, we know that for every sampled $x$, there must be one corresponding class (implied by the separability condition), denoted by $y_x$. Moreover, by weak realizability, we have $\bm{p}_{\theta^*}(x) = \bm{p}(x)$. Therefore,
\begin{equation*}
    \begin{aligned}
        g_{ca,\theta^*}(x) &= \sum_{k=1}^K \ell(\hat{f}(x), k)p_{\theta^*}(x)_k = \ell(\hat{f}(x), y_x), \quad \text{and}\\
        g(x) &= \sum_{k=1}^K \ell(\hat{f}(x), k)p(x)_k = \ell(\hat{f}(x), y_x).
    \end{aligned}
\end{equation*}
Notice that this is independent of $\hat{f}$, and hence $g(x) = g_{ca,\theta^*}(x)$ for all $\hat{f}$. Therefore, for the population risk,
\begin{equation*}
    \begin{aligned}
        R_{ca}(\theta^*)
        &= \mathbb{E}[(g_{ca,\theta^*}(X) - g(X))^2] \\
        &= \int (\ell(\hat{f}(x), y_x) - \ell(\hat{f}(x), y_x))^2 p(x)\,dx = 0.
    \end{aligned}
\end{equation*}
The first bound then follows from \eqref{eqn_radmacher_classification}.

For the regression-based method, it suffices to prove the existence of a predictor $\hat{f}$ such that for its corresponding conditional loss
$g(x) = \mathbb{E}[\ell(\hat{f}(x), Y)\mid X = x],$
we have
$$
R(\theta^*) = \inf_{\theta\in\Theta} R(\theta) = \inf_{\theta\in\Theta}\mathbb{E}[|g(X) - g_{\theta}(X)|] > 0.
$$
Without loss of generality, we can prove this by considering a binary classification problem in which $\hat{f}(x)$ is the indicator function of the \textit{fat Cantor set} $\mathbb{C}$ \citep{aliprantis1998principles}, that is,
$$
\hat{f}(x) =
\begin{cases}
    1 & \text{if } x \in \mathbb{C},\\
    0 & \text{if } x \notin \mathbb{C}.
\end{cases}
$$
Note that $\mathbb{C}$ is uncountable, nowhere dense, and has positive measure. Therefore, because the data are continuously distributed, there exists a region $D$ with positive support such that $y_x = 1$ for all $x \in D$. Hence,
$$
g(x) = \mathbb{E}[\ell(\hat{f}(x), Y)\mid X = x] = \sum_{k=0}^1 \ell(\hat{f}(x), k)p_k(x) = \ell(\hat{f}(x), y_x),
$$
which implies that for $x \in D$,
$$
g(x) =
\begin{cases}
    0 & \text{if } x \in \mathbb{C},\\
    1 & \text{if } x \notin \mathbb{C}.
\end{cases}
$$
Notice that the function $g(x)$ does not belong to any continuous hypothesis class $\mathcal{G}$. Therefore, there exists $\bar{\delta} > 0$ such that
$$
\inf_{\theta \in \Theta} \int_D |g(x) - g_{\theta}(x)|\,dx \geq \bar{\delta}.
$$
Since $D$ has positive support, there exists $\delta > 0$ such that
$$
R(\theta^*) = \inf_{\theta\in\Theta} \mathbb{E}[|g(X) - g_{\theta}(X)|]
\geq
\inf_{\theta\in\Theta} \mathbb{E}[|g(X) - g_{\theta}(X)| \cdot 1_D]
> \delta.
$$
Combining this with \eqref{eqn_radmacher_general} completes the proof of the second part.

\section{Classification Experiment Details}

For $\hat{f}_1$, a CNN-based classifier, we employ a minimal convolutional neural network for CIFAR-10 classification. The network consists of two convolutional blocks followed by two fully connected layers. In the first block, a 3×3 convolution with 16 filters (stride 1, padding 1) is applied to the 32×32×3 input, followed by a ReLU activation and a 2×2 max pooling operation that halves the spatial resolution to 16×16. In the second block, a 3×3 convolution with 32 filters (stride 1, padding 1) is used, again followed by ReLU and a 2×2 max pooling, reducing the spatial dimensions to 8×8. The resulting feature map (of size 32×8×8) is flattened and fed into a fully connected layer with 64 units (with ReLU activation), followed by a final fully connected layer that produces logits for 10 classes.

For $\hat{f}_2$, a ResNet-based classifier, we utilize the ResNet‑8 model, which is a compact residual network designed for CIFAR‑10 classification. It begins with a 3×3 convolution that maps the 3-channel input to 16 feature maps, followed by batch normalization and a ReLU activation. This is followed by three residual layers, each consisting of a single basic block. Each basic block contains two 3×3 convolutional layers (with batch normalization and ReLU activations) and a shortcut connection that either passes the input unchanged or applies a 1×1 convolution (with batch normalization) to match dimensions when necessary. In our configuration, the first block preserves the spatial resolution, while the second and third blocks use a stride of 2 in their first convolution to downsample the feature maps. Finally, a global average pooling layer aggregates the spatial information, and a fully connected layer produces the final class scores. The network comprises eight layers: one initial convolution, six convolutional layers within the residual blocks, and one final fully connected layer.

For $\hat{f}_3$, an EfficientNet-based classifier, we initialize an EfficientNet‑B0 model pre‑trained on ImageNet (using the EfficientNet\_B0\_Weights.IMAGENET1K\_V1). The default classifier in EfficientNet‑B0 is a sequential block containing a dropout layer followed by a linear layer that maps a 1280-dimensional feature vector to 1000 classes. To adapt the model for CIFAR‑10, we replace the final linear layer with a new one that outputs 10 classes. 

In order to use representations as input for regressing conditional risk using a computer vision regressor (say CNN), we extract representations from a computer vision classifier (say EfficientNet) by removing its final linear layer, yielding a 1280-dimensional embedding, and train a simple multilayer perceptron (with one hidden layer of 256 units) to regress the cross-entropy loss of the CNN prediction model. The training loss function is mean squared error. We then use this calibrated loss to guide our decisions in the L2D setting.

We partition the CIFAR‑10 dataset into three segments: 50\% of the data is used to train the classification predictors, 40\% is dedicated to training the feature representation and the MLP loss predictor, and the remaining 10\% is reserved for testing the L2D loss. Standard data augmentations -- including random cropping, horizontal flipping, and mean normalization -- are applied during training, which is conducted using an early stopping strategy with a five-round patience. The predictors are optimized using cross-entropy loss and the AdamW optimizer, with a learning rate of $1\times10^{-3}$ and a weight decay of $1\times10^{-4}$. The training accuracies of the three predictors $\hat{f}_1$, $\hat{f}_2$, and $\hat{f}_3$ are 77.68\%, 85.44\%, and 92.39\%, respectively.

\section{Regression With Rejection Experiment Details}
\label{app_RwR_results}

This section presents the full results of the RwR experiments along with implementation details.

\subsection{Data Description}
\label{app_RwR_results_data}

For this experiment, we used eight regression datasets from the UCI Machine Learning Repository \citep{uci_datasets_ref}. The datasets were accessed via the \texttt{uci\_datasets} Python package, which provides preprocessed datasets with predefined train-test splits for benchmarking studies (\url{https://github.com/treforevans/uci_datasets}). A brief description of each dataset is provided below:
\begin{itemize}
    \item \textbf{Concrete}: Predicts the compressive strength of concrete based on its ingredients and age. Contains 1030 instances with 8 features.  
    \item \textbf{Wine}: Predicts wine quality based on physicochemical tests. The red wine subset contains 1599 instances with 11 features.  
    \item \textbf{Airfoil}: Predicts the scaled sound pressure level of airfoil blade sections tested in an anechoic wind tunnel. Contains 1503 instances with 5 features.  
    \item \textbf{Energy}: Predicts the energy efficiency of buildings based on building parameters. Contains 768 instances with 8 features.  
    \item \textbf{Housing}: Predicts the median home value in Boston based on various housing features. Contains 506 instances with 13 features.  
    \item \textbf{Solar}: Predicts the number of X-class solar flares in a 24-hour period. Contains 1066 instances with 10 features.  
    \item \textbf{Forest}: Predicts the burned area of forest fires based on environmental factors. Contains 517 instances with 12 features.  
    \item \textbf{Parkinsons}: Predicts UPDRS scores from individuals with early-stage Parkinson's disease. Contains 5875 instances with 20 features.  
\end{itemize}

\subsection{Implementation Details}
\label{app_RwR_results_implementation}

In this experiment, we trained 16 regressor-calibrator pairs for each dataset. Each model in a pair was selected from one of the following options:

\begin{itemize} 
\item \textbf{LR}: Standard linear regression. 
\item \textbf{RF}: Random forest with no maximum depth limit. 
\item \textbf{MLP}: A multilayer perceptron with a single hidden layer containing 64 units. This model uses ReLU activation and the Adam optimizer. Training was performed with a batch size of 256, an initial learning rate of 0.0005, and a maximum of 800 iterations. 
\item \textbf{MLP2}: A multilayer perceptron with two hidden layers, each containing 64 units. Like the \textbf{MLP} model, it uses ReLU activation and the Adam optimizer, with the same training parameters (batch size of 256, initial learning rate of 0.0005, and a maximum of 800 iterations). 
\end{itemize}

All models were implemented using the \texttt{scikit-learn} Python package (v1.2.1). Feature standardization was applied to all models -- except for Random Forest -- using \texttt{StandardScaler()} from \texttt{scikit-learn}, which removes the mean and scales features to unit variance. The models were trained using the squared loss objective function. To ensure reproducibility, we fixed the random seed at 42 for all algorithms.

We performed 10-fold cross-validation using splits provided by the \texttt{uci\_datasets} package. In each fold, the dataset was partitioned into three subsets: 50\% of the data were used to train the regressor $\hat{f}$, 40\% to train the calibrator $\hat{R}$, and the remaining 10\% were reserved for testing. The results reported for this experiment are the averages across the 10 folds.

\subsection{Full Regression with Rejection Results}
\label{app_RwR_results_full_table}

\begin{table}[t]
\centering
\caption{Full RwR Experiment Results.}
\label{comp_performance_full}
\vspace{0.5em}
\resizebox{1.0\columnwidth}{!}{
\setlength\extrarowheight{5pt}
\begin{tabular}{cc|cccc|cccc|cccc|cccc|cc}
\textbf{Data} & \multicolumn{1}{c|}{\textbf{Cost}} 
& \multicolumn{1}{c}{\textbf{\begin{tabular}[c]{@{}c@{}}LR\\ \noalign{\vskip -4pt} +\\ \noalign{\vskip -4pt} LR\end{tabular}}}
& \multicolumn{1}{c}{\textbf{\begin{tabular}[c]{@{}c@{}}LR\\ \noalign{\vskip -4pt} +\\ \noalign{\vskip -4pt} MLP\end{tabular}}} 
& \multicolumn{1}{c}{\textbf{\begin{tabular}[c]{@{}c@{}}LR\\ \noalign{\vskip -4pt} +\\ \noalign{\vskip -4pt} MLP2\end{tabular}}} 
& \multicolumn{1}{c|}{\textbf{\begin{tabular}[c]{@{}c@{}}LR\\ \noalign{\vskip -4pt} +\\ \noalign{\vskip -4pt} RF\end{tabular}}}
& \multicolumn{1}{c}{\textbf{\begin{tabular}[c]{@{}c@{}}MLP\\ \noalign{\vskip -4pt} +\\ \noalign{\vskip -4pt} LR\end{tabular}}} 
& \multicolumn{1}{c}{\textbf{\begin{tabular}[c]{@{}c@{}}MLP\\ \noalign{\vskip -4pt} +\\ \noalign{\vskip -4pt} MLP\end{tabular}}} 
& \multicolumn{1}{c}{\textbf{\begin{tabular}[c]{@{}c@{}}MLP\\ \noalign{\vskip -4pt} +\\ \noalign{\vskip -4pt} MLP2\end{tabular}}} 
& \multicolumn{1}{c|}{\textbf{\begin{tabular}[c]{@{}c@{}}MLP\\ \noalign{\vskip -4pt} +\\ \noalign{\vskip -4pt} RF\end{tabular}}}
& \multicolumn{1}{c}{\textbf{\begin{tabular}[c]{@{}c@{}}MLP2\\ \noalign{\vskip -4pt} +\\ \noalign{\vskip -4pt} LR\end{tabular}}} 
& \multicolumn{1}{c}{\textbf{\begin{tabular}[c]{@{}c@{}}MLP2\\ \noalign{\vskip -4pt} +\\ \noalign{\vskip -4pt} MLP\end{tabular}}} 
& \multicolumn{1}{c}{\textbf{\begin{tabular}[c]{@{}c@{}}MLP2\\ \noalign{\vskip -4pt} +\\ \noalign{\vskip -4pt} MLP2\end{tabular}}} 
& \multicolumn{1}{c|}{\textbf{\begin{tabular}[c]{@{}c@{}}MLP2\\ \noalign{\vskip -4pt} +\\ \noalign{\vskip -4pt} RF\end{tabular}}}
& \multicolumn{1}{c}{\textbf{\begin{tabular}[c]{@{}c@{}}RF\\ \noalign{\vskip -4pt} +\\ \noalign{\vskip -4pt} LR\end{tabular}}} 
& \multicolumn{1}{c}{\textbf{\begin{tabular}[c]{@{}c@{}}RF\\ \noalign{\vskip -4pt} +\\ \noalign{\vskip -4pt} MLP\end{tabular}}} 
& \multicolumn{1}{c}{\textbf{\begin{tabular}[c]{@{}c@{}}RF\\ \noalign{\vskip -4pt} +\\ \noalign{\vskip -4pt} MLP2\end{tabular}}} 
& \multicolumn{1}{c|}{\textbf{\begin{tabular}[c]{@{}c@{}}RF\\ \noalign{\vskip -4pt} +\\ \noalign{\vskip -4pt} RF\end{tabular}}}
& \multicolumn{1}{c}{\textbf{\begin{tabular}[c]{@{}c@{}}SelNet\end{tabular}}}
&  \multicolumn{1}{c}{\textbf{\begin{tabular}[c]{@{}c@{}}NN\\ \noalign{\vskip -4pt} +\\ \noalign{\vskip -4pt} kNNRej\end{tabular}}} \\ 
\hline

\multirow{4}{*}{\rotatebox{90}{\textbf{Concrete}}}
& 0.2 & 1.31 & \textbf{0.20} & \textbf{0.20} & \textbf{0.20} & 1.32 & \textbf{0.20} & 0.23 & \textbf{0.20} & 1.15 & \textbf{0.20} & 0.93 & \textbf{0.20} & 1.42 & \textbf{0.20} & 1.37 & \textbf{0.20} & 0.75 & \textbf{0.20} \\
& 0.5 & 1.61 & \textbf{0.50} & \textbf{0.50} & \textbf{0.50} & 1.61 & \textbf{0.50} & 0.53 & \textbf{0.50} & 1.45 & \textbf{0.50} & 1.24 & \textbf{0.50} & 1.71 & \textbf{0.50} & 1.72 & \textbf{0.50} & 1.09 & \textbf{0.50} \\
& 1 & 2.10 & \textbf{1.00} & \textbf{1.00} & \textbf{1.00} & 2.09 & \textbf{1.00} & 1.02 & \textbf{1.00} & 2.19 & \textbf{1.00} & 1.73 & 1.02 & 2.21 & \textbf{1.00} & 2.21 & \textbf{1.00} & 2.82 & \textbf{1.00} \\
& 2 & 3.08 & \textbf{2.00} & \textbf{2.00} & \textbf{2.00} & 3.08 & \textbf{2.00} & 2.02 & \textbf{2.00} & 3.24 & \textbf{2.00} & 2.81 & 2.01 & 3.87 & 2.03 & 3.27 & \textbf{2.00} & 3.16 & \textbf{2.00} \\ 
\hline

\multirow{4}{*}{\rotatebox{90}{\textbf{Wine}}}
& 0.2 & 0.22 & 0.21 & 0.21 & 0.19 & 0.19 & 0.19 & 0.18 & \textbf{0.17} & 0.20 & 0.20 & 0.19 & \textbf{0.17} & 0.21 & 0.21 & 0.20 & 0.18 & 0.18 & 0.23 \\
& 0.5 & 0.33 & 0.33 & 0.33 & 0.31 & 0.25 & 0.25 & \textbf{0.24} & \textbf{0.24} & 0.28 & 0.27 & 0.26 & 0.25 & 0.29 & 0.29 & 0.28 & 0.28 & \textbf{0.24} & 0.27 \\
& 1 & 0.36 & 0.35 & 0.37 & 0.34 & \textbf{0.25} & \textbf{0.25} & \textbf{0.25} & \textbf{0.25} & 0.27 & 0.27 & 0.28 & 0.27 & 0.30 & 0.29 & 0.30 & 0.30 & \textbf{0.25} & 0.28 \\
& 2 & 0.36 & 0.36 & 0.36 & 0.36 & \textbf{0.25} & \textbf{0.25} & \textbf{0.25} & \textbf{0.25} & 0.26 & 0.27 & 0.27 & 0.26 & 0.30 & 0.29 & 0.30 & 0.31 & \textbf{0.25} & 0.28 \\
\hline

\multirow{4}{*}{\rotatebox{90}{\textbf{Airfoil}}}
& 0.2 & \textbf{0.20} & 0.26 & 0.68 & \textbf{0.20} & 0.69 & 0.30 & 0.38 & \textbf{0.20} & 0.36 & 0.25 & 0.50 & \textbf{0.20} & 0.31 & 0.29 & 0.37 & \textbf{0.20} & 0.21 & \textbf{0.20} \\
& 0.5 & 0.50 & 0.56 & 0.97 & 0.50 & 0.98 & 0.60 & 0.73 & 0.50 & 0.70 & 0.56 & 0.82 & 0.50 & 0.62 & 0.59 & 0.68 & 0.50 & \textbf{0.49} & 0.50 \\
& 1 & 1.00 & 1.06 & 1.45 & 1.00 & 1.47 & 1.11 & 1.25 & 0.99 & 1.19 & 1.05 & 1.26 & \textbf{0.96} & 1.09 & 1.08 & 1.14 & 0.97 & 0.99 & 1.00 \\
& 2 & 2.06 & 2.08 & 2.41 & 1.94 & 2.43 & 2.11 & 2.21 & 1.91 & 2.27 & 1.91 & 2.15 & 1.84 & 1.98 & 1.99 & 1.99 & \textbf{1.73} & 1.84 & 2.02 \\
\hline

\multirow{4}{*}{\rotatebox{90}{\textbf{Energy}}}
& 0.2 & 0.77 & 0.52 & 0.31 & 0.26 & 0.44 & 0.43 & 0.27 & 0.20 & 0.23 & 0.23 & 0.27 & 0.20 & 0.17 & 0.16 & 0.17 & \textbf{0.14} & 0.26 & 0.21 \\
& 0.5 & 1.03 & 0.79 & 0.61 & 0.54 & 0.69 & 0.69 & 0.56 & 0.46 & 0.49 & 0.49 & 0.50 & 0.45 & 0.28 & 0.27 & 0.28 & \textbf{0.26} & 0.49 & 0.58 \\
& 1 & 1.46 & 1.20 & 1.03 & 0.85 & 1.06 & 1.18 & 0.98 & 0.80 & 0.79 & 0.77 & 0.78 & 0.71 & 0.31 & 0.32 & 0.32 & \textbf{0.30} & 0.86 & 1.03 \\
& 2 & 2.17 & 1.93 & 1.62 & 1.29 & 1.65 & 1.97 & 1.60 & 1.32 & 1.18 & 1.15 & 1.13 & 1.04 & 0.33 & \textbf{0.32} & 0.34 & \textbf{0.32} & 1.39 & 1.60 \\
\hline

\multirow{4}{*}{\rotatebox{90}{\textbf{Housing}}}
& 0.2 & 2.94 & 0.23 & 1.39 & \textbf{0.20} & 2.17 & 0.48 & 1.60 & \textbf{0.20} & 1.82 & 0.48 & 1.38 & \textbf{0.20} & 1.58 & 0.47 & 1.02 & 
\textbf{0.20} & 1.27 & 0.22 \\
& 0.5 & 3.19 & 0.53 & 1.70 & \textbf{0.50} & 2.41 & 0.84 & 1.84 & \textbf{0.50} & 2.13 & 0.78 & 1.65 & \textbf{0.50} & 1.84 & 0.75 & 1.33 & \textbf{0.50} & 1.31 & 0.59 \\
& 1 & 3.62 & 1.03 & 2.20 & \textbf{1.00} & 2.97 & 1.33 & 2.46 & \textbf{1.00} & 2.62 & 1.26 & 2.12 & \textbf{1.00} & 2.27 & 1.34 & 1.81 & 1.01 & 2.23 & \textbf{1.00} \\
& 2 & 4.64 & 2.44 & 3.27 & 2.04 & 3.95 & 2.33 & 3.28 & \textbf{2.00} & 3.49 & 2.38 & 3.19 & 2.01 & 3.18 & 2.31 & 2.73 & 2.07 & 2.47 & \textbf{2.00} \\
\hline

\multirow{4}{*}{\rotatebox{90}{\textbf{Solar}}}
& 0.2 & 0.19 & 0.23 & 0.35 & 0.24 & 0.20 & 0.23 & 0.30 & 0.23 & 0.20 & 0.25 & 0.38 & 0.24 & 0.19 & 0.24 & 0.33 & 0.23 & \textbf{0.18} & 0.19 \\
& 0.5 & 0.37 & 0.39 & 0.54 & 0.41 & 0.37 & 0.36 & 0.46 & 0.40 & 0.37 & 0.39 & 0.54 & 0.44 & \textbf{0.35} & 0.36 & 0.50 & 0.41 & 0.45 & 0.40 \\
& 1 & \textbf{0.50} & 0.53 & 0.62 & 0.54 & 0.51 & 0.54 & 0.64 & 0.54 & 0.52 & 0.57 & 0.63 & 0.55 & 0.55 & 0.59 & 0.70 & 0.57 & 0.52 & 0.60 \\
& 2 & \textbf{0.60} & 0.68 & 0.66 & 0.63 & 0.63 & 0.70 & 0.68 & 0.64 & 0.66 & 0.73 & 0.72 & 0.66 & 0.71 & 0.72 & 0.81 & 0.73 & 0.68 & 0.75 \\
\hline

\multirow{4}{*}{\rotatebox{90}{\textbf{Forest}}}
& 0.2 & 0.30 & 0.33 & 0.52 & \textbf{0.20} & 0.28 & 0.32 & 0.56 & \textbf{0.20} & 0.32 & 0.33 & 0.61 & \textbf{0.20} & 0.23 & 0.28 & 0.48 & \textbf{0.20} & 0.54 & 0.30 \\
& 0.5 & 0.60 & 0.65 & 0.83 & 0.51 & 0.58 & 0.65 & 0.90 & 0.51 & 0.63 & 0.67 & 0.87 & 0.51 & 0.54 & 0.59 & 0.82 & 0.51 & 0.78 & \textbf{0.50} \\
& 1 & 1.13 & 1.22 & 1.30 & 1.07 & 1.13 & 1.17 & 1.33 & 1.08 & 1.14 & 1.16 & 1.33 & 1.05 & 1.06 & 1.08 & 1.39 & \textbf{1.04} & 1.25 & 1.09 \\
& 2 & 2.03 & 1.96 & 1.91 & 2.00 & 2.00 & 1.95 & \textbf{1.85} & 1.97 & 2.03 & 1.98 & 1.89 & 1.97 & 1.95 & 1.96 & 2.06 & 1.98 & 2.17 & 2.13 \\
\hline

\multirow{4}{*}{\rotatebox{90}{\textbf{Parkinsons}}}
& 0.2 & 0.78 & 0.66 & 1.31 & 0.20 & 0.72 & 0.62 & 1.12 & 0.20 & 0.62 & 0.52 & 0.80 & 0.20 & 0.19 & 0.17 & 0.19 & \textbf{0.12} & 0.20 & 0.20 \\
& 0.5 & 1.08 & 0.96 & 1.60 & 0.50 & 1.02 & 0.94 & 1.41 & 0.50 & 0.91 & 0.83 & 1.07 & 0.50 & 0.33 & 0.28 & 0.29 & \textbf{0.20} & 0.60 & 0.50 \\
& 1 & 1.57 & 1.47 & 2.08 & 1.00 & 1.53 & 1.44 & 1.90 & 1.00 & 1.36 & 1.30 & 1.51 & 1.00 & 0.37 & 0.34 & 0.33 & \textbf{0.26} & 1.17 & 1.00 \\
& 2 & 2.57 & 2.48 & 3.02 & 2.00 & 2.58 & 2.52 & 2.89 & 2.00 & 2.26 & 2.15 & 2.33 & 1.89 & 0.35 & 0.37 & 0.33 & \textbf{0.32} & 1.97 & 2.00 \\

\end{tabular}
}
\end{table}

Table~\ref{comp_performance_full} presents the RwR losses for all 16 regressor-calibrator pairs, along with two benchmark RwR algorithms: SelectiveNet \citep{pmlr-v97-geifman19a} and NN+kNNRej \citep{pmlr-v238-li24g}. SelectiveNet is a deep learning approach with an integrated rejector, whereas NN+kNNRej is a no-rejection learning algorithm that achieved the best performance in the study by \cite{pmlr-v238-li24g}. We used the publicly available implementations of these algorithms from \url{https://github.com/hanzhao-wang/RwR}.

\end{document}